\documentclass[nohyperref]{article}

\usepackage{color,soul}

\usepackage{microtype}
\usepackage{graphicx}
\usepackage{subfigure}
\usepackage{booktabs} 

\usepackage{hyperref}

\usepackage[accepted]{icml2022}

\usepackage{amsmath}
\usepackage{amssymb}
\usepackage{mathtools}
\usepackage{amsthm}

\usepackage[capitalize,noabbrev]{cleveref}

\theoremstyle{plain}

\theoremstyle{definition}

\theoremstyle{remark}

\usepackage{amsmath}
\usepackage{tcmmacros}
\usepackage[textsize=tiny]{todonotes}

\icmltitlerunning{SITHCon: A scale-invariant temporal convolutional network}
\begin{document}
\twocolumn[
\icmltitle{A deep convolutional neural network that is invariant to time rescaling}
\begin{icmlauthorlist}
\icmlauthor{Brandon G. Jacques}{uva}
\icmlauthor{Zoran Tiganj}{indiana}
\icmlauthor{Aakash Sarkar}{boston}
\icmlauthor{Marc W. Howard}{boston}
\icmlauthor{Per B. Sederberg}{uva}
\end{icmlauthorlist}

\icmlaffiliation{uva}{Department of Psychology, University of Virginia, Charlottesville, VA, United States}
\icmlaffiliation{indiana}{Department of Computer Science, Indiana University, Bloomington, IN, United States}
\icmlaffiliation{boston}{Department of Psychological and Brain Sciences, Boston University, Boston, MA, United States}

\icmlcorrespondingauthor{Per B. Sederberg}{pbs5u@virginia.edu}
\icmlkeywords{Neurally-inspired deep networks; time cells; scale-invariant
networks} 
\vskip 0.3in
]
\printAffiliationsAndNotice{
This material is based upon work supported by the Defense Advanced Research
Projects Agency (DARPA) under Agreement No.~HR00112190036, ONR MURI N00014-16-1-2832, and a Google-AI
Faculty Research Award. The authors acknowledge Research computing at The
University of Virginia for providing computational resources and technical
support that have contributed to the results reported within this publication.
URL: https://rc.virginia.edu.
} 

\begin{abstract}
Human learners can readily understand speech, or a melody, when it is
presented slower or faster than usual. This paper presents a deep
Scale-Invariant Temporal History Convolution network (SITHCon) that uses a
logarithmically compressed temporal representation at each level.  Because time
rescaling of the input results in a translation
of the memory representation over $\log$ time, and because the output of the
convolution is equivariant to translations, this network can generalize to
out-of-sample data that are temporal rescalings of a learned pattern.  We
compare the performance of SITHCon to a Temporal Convolution Network (TCN) on
classification and regression problems with both univariate and multivariate
time series.   We find that SITHCon, unlike TCN, generalizes robustly over
rescalings of about an order of magnitude.  Moreover, we show that the network
can generalize over exponentially large scales without retraining the weights
simply by extending the range of the logarithmically-compressed temporal
memory.  \end{abstract}

\section{Introduction}
\label{introduction}

Many problems in machine perception require integration of information
over continuous time.  In the natural world, these temporal signals
can unfold over different time scales.  For instance, one would want a
speech recognition system to be able to identify words spoken more
quickly than usual---perhaps the user is in a hurry---as well as more
slowly---perhaps the user is tired or has one of several neurological
conditions that affect the rate of speech.  People can naturally adapt
to time series presented at different rates.  For instance most people
can easily identify a familiar melody played at an unfamiliar speed,
yet this is a class of problems that proven difficult for many forms
of AI. Although deep neural networks have revolutionized a number of
fields that rely on representing time series, including speech
perception \citep{Lea.etal.2017}, they do not generalize across rates
of presentation and need to be explicitly trained on a wide range of
time scales \cite{Chan.etal.2021}.  This paper presents a deep
convolutional neural network (CNN), inspired by recent work in
neuroscience, that generalizes to time series presented at untrained
rates.

\begin{figure}[h!]
	\centering
	\includegraphics[width=0.95\columnwidth]{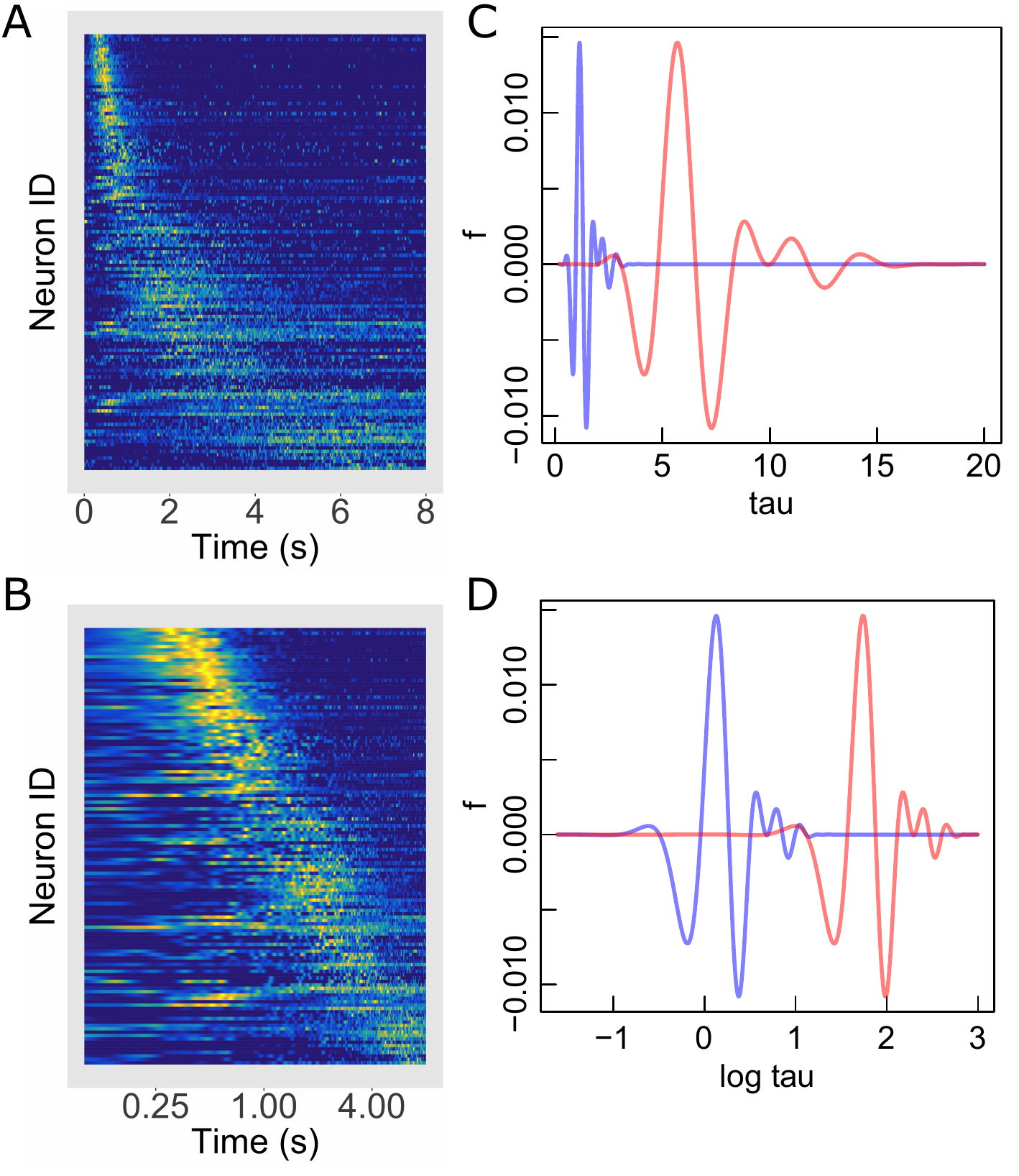}
	\caption{
	The brain represents a logarithmically-compressed temporal memory that
	turns rescaling of physical time into translation of the memory.
	\textbf{A.} Time cells in the rodent hippocampus fire in the time
	after a relevant stimulus.  Each line shows the firing rate of one
	neuron during time within the delay period of a memory experiment.
	The cells are sorted on their peak time.  Note the curvature of the
	central ridge.  \textbf{B.} The same data shown as a function of
	$\log$ time.  Note that the central ridge appears as a straight line
	with a constant width when plotted as a function of $\log t$.  After
	Cao, et al., (2021).  \textbf{C.}    A function, $f(t)$ and a time
	rescaled version $f(5 t)$.  \textbf{D.}  The
	same functions plotted as a function of $\log t$.  Note that rescaling
	stretches the functions as a function of time but
	results in a translation as a function of  $\log$ time.
		\label{fig:logtimecells}
	}
\end{figure}

The way the mammalian brain retains information about the time of past events
provides a novel strategy to construct deep networks that are invariant to
rescalings of their inputs.  Populations of neurons referred to as ``time cells''
(within the hippocampus, entorhinal cortex, and lateral prefrontal cortex) fire
in sequence after a triggering  stimulus \citep[Fig.~\ref{fig:logtimecells},][]{Eichenbaum.2014}.
Different time cells fire at different characteristic times after the
triggering stimulus form a temporal basis set. Because different
sequences of cells are triggered by different environmental stimuli
\citep[e.g.,][]{Tiganj.etal.2018, Taxidis.etal.2020}, the population forms a
representation of \emph{what happened when} in the past.  Critically, the
temporal basis set is compressed \cite{Kraus.etal.2013}.  Psychological data
and theoretical considerations suggest that the basis set ought to evenly
cover $\log$ time rather than linear time
\cite{Balsam.Gallistel.2009,ShanHowa13,Tiganj.etal.2018, Howard.etal.2015}.
Notably, neurophysiological evidence suggests that the brain uses a
logarithmically-compressed temporal memory in a number of widely-spaced brain
regions, including auditory cortex, cerebellum and hippocampus
\cite{RahmEtal20, Guo.etal.2020, CaoEtal21}.

A temporal memory constructed over $\log$ time is uniquely robust to temporal
rescalings.  Consider a time series that is rescaled by a factor $a$, $t
\rightarrow at$ (Fig.~\ref{fig:logtimecells}).  Because  \(\log (at) = \log(t)
+ \log(a)\), rescaling time by a factor \(a\) simply results in a translation,
by a factor \(\log(a)\), along a logarithmically-compressed temporal memory.
In computer vision deep CNNs have been
extremely successful because they are equivariant, modulo edge effects, to
translation of their input.  By building a CNN with a maxpool operation over a
logarithmically-compressed temporal memory, we construct a network, referred
to as SITHCon  whose output is invariant to time rescaling.
(Fig.~\ref{sithcon_network}).  We contrast SITHCon with a
Temporal Convolution Network (TCN).  TCNs are deep
CNNs constructed over a linear
temporal memory that have been applied, often with state of the art
performance, to speech recognition, sequence modeling, and action recognition
from video \cite{Bai.etal.2018, Lea.etal.2017}. Because the TCN uses a
standard  temporal memory that samples the input signal at evenly-spaced time
points, time rescaling of the input does not result in translation of the
memory.  As a consequence, the TCN should not generalize over temporal
rescaling.

\section{Methods}
\label{methods}
Each layer of SITHCon is composed of a
logarithmically-compressed temporal memory---SITH---followed by a
convolutional layer and a dense layer.
The logarithmically-compressed memory is the primary novel component of the
network and is responsible for the generalization to rescaled inputs.

\subsection{Scale Invariant Temporal History (SITH)}
\label{scale-invariant-temporal-history-sith}

\newcommand{\scale}[1]{\ensuremath{\mathcal{S}_{#1}}}
\newcommand{\trans}[1]{\ensuremath{\mathcal{T}_{#1}}}

\newcommand{\ftildevec}{\ensuremath{\mathbf{\ftilde}}}
\newcommand{\ftildefeat}{\ensuremath{\Phi_k}}

\newcommand{\ntaustar}{\ensuremath{N_{\taustar}}}
\newcommand{\taustarmin}{\ensuremath{\taustar_{1}}}
\newcommand{\nfeatf}{\ensuremath{N_{f}}}
\newcommand{\nfeat}{\ensuremath{N}}

\newcommand{\convwidth}{\ensuremath{K}}
\newcommand{\convfilt}{\ensuremath{g}}

\newcommand{\layer}[1]{\ensuremath{^{\left(#1\right)}}}

The goal of the scale-invariant temporal memory is to provide a record of the
recent past as a function of time at each moment.
Given an input signal \(f(t)\), let us define the history leading up to
the present time $t$ as \(f_t(\tau)=f(t-\tau)\), where \(\tau\) runs from zero at
the present to infinity in the remote distant past.
The temporal memory estimates $f_t(\tau)$  in the neighborhood of $\ntaustar$
discrete time points $\taustar_n$.  We refer to the state of the memory at
time $t$ as $\ftilde_t[n]$.

Two properties enable the SITH buffer to be logarithmically-compressed.
First, rather than choosing the difference between adjacent values of
$\taustar$ to be constant, the \emph{ratio} between adjacent values is
constant:
\begin{equation}
        \taustar_n = \left(1+c\right)^{n-1} \taustarmin,
        \label{eq:geometric}
\end{equation}
where $c$ is positive, and derived from the parameters $\tau_{min}$,
$\tau_{max}$, and $N$:

\begin{equation}
    c = \frac{\tau_{max}}{\tau_{min}}^{\frac{1}{N - 1}} - 1.
\end{equation}

Equation~\ref{eq:geometric} implies that the temporal
receptive fields are evenly separated as a function of $\log$ time
\(\log{\taustar_{n+1}} - \log{\taustar_n} = 1+c\).
Second, the temporal receptive field of each node is a function of
$\tau/\taustar_n$:
\begin{eqnarray}
	\nonumber\ftilde_t[n] &=&
            \int_{0}^{\infty}
                \phi\left(
                        \frac{\tau}{\taustar_n}
                    \right) f_t(\tau)
            \ d\tau \\
	   & = &
            \int_0^{\infty}
\phi(\tau')\ f_t(\taustar_n \tau')
     d\tau'\\
     & = & \ftildefeat \circ f_t(\tau).
        \label{eq:conveqs}
\end{eqnarray}

\begin{figure}[t]
\centering
\includegraphics[width=0.9\columnwidth]{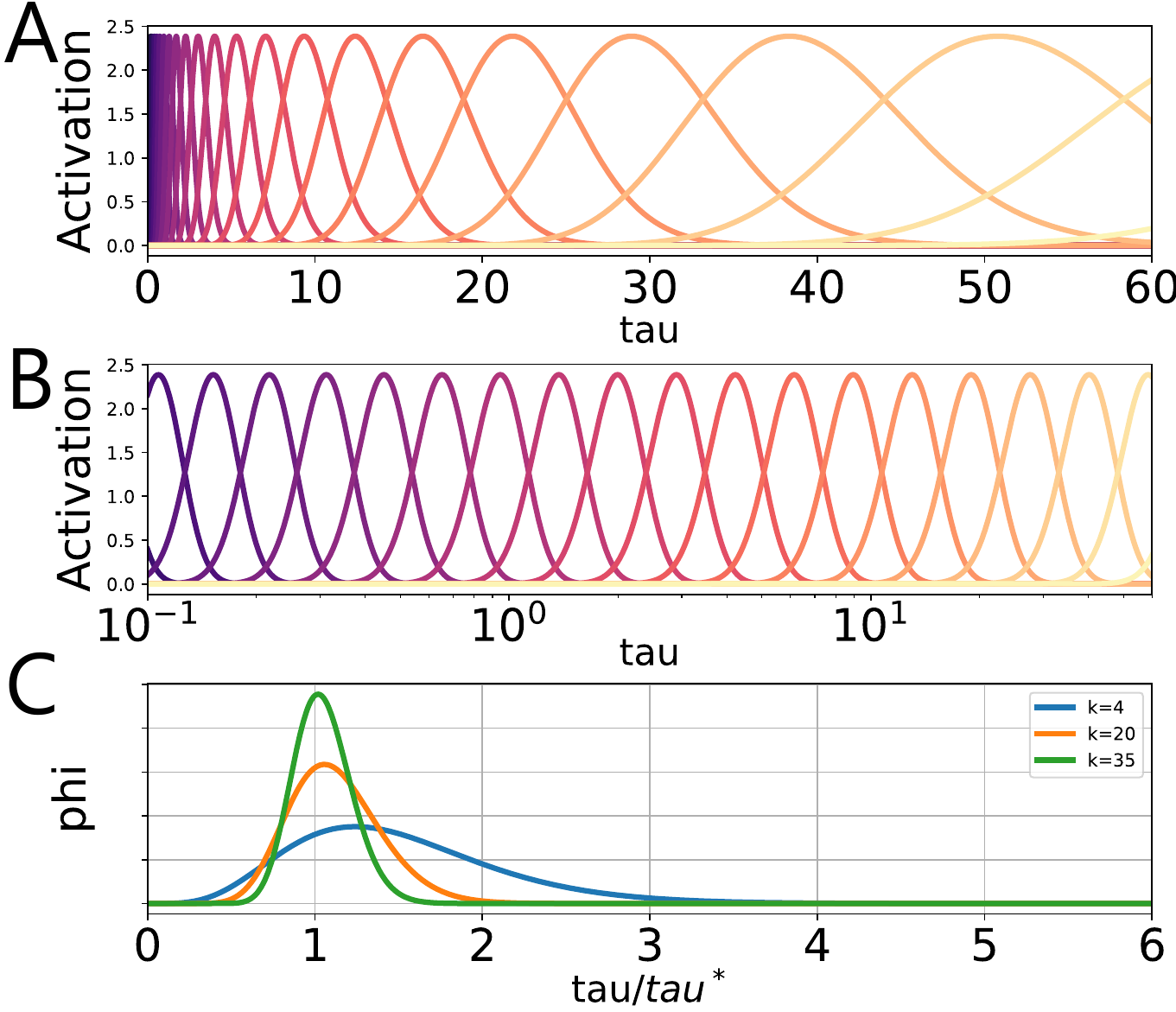}
\caption{\emph{The effect of k on the temporal receptive fields in SITH.}
	\textbf{A.} Plot of receptive fields $\phi(\tau/\taustar_n)$ for
	$\taustar$'s in geometric series.  For clarity, each receptive field
	has been scaled to have the same peak (\emph{i.e.,} multiplied by
	$\taustar_n$).
	\textbf{B.}  The same receptive fields plotted on a $\log$ scale.
	\textbf{C.}  The function $\phi(\tau/\taustar)$ for different values
	of $k$.
	Larger $k$ results in tighter receptive fields.}
	\label{filters}
\end{figure}

The particular choice of $\phi$ fixes the shape of the receptive fields.  Here
we choose $\phi(x) = x^k \exp(-kx)$ for some constant $k$. The effects of $k$
on the shape of the receptive fields can be seen in Fig. \ref{filters}.
$\phi$ is a unimodal
function that peaks at 1.
As $k$ becomes larger, the function $\phi$ becomes more sharply peaked.
Because each temporal receptive field has the same shape relative to
$\taustar_n$ and because the $\taustar_n$ are evenly-spaced as a function of
$\log \tau$, the temporal memory evenly samples $f_t(\tau)$ as a function of
$\log \tau$.

\subsubsection{Rescaling time induces a translation of activity in SITH}
\label{sec:scaletrans}
In this study we are interested in the effects of time-rescaling the input
\(\tau \rightarrow a \tau\).  Equation~\ref{eq:conveqs} makes
clear that, for a particular node $n$,  rescaling $\tau$ can be undone by taking
\(\taustar_n \rightarrow \taustar_n/a\).  However, because the $\taustar$s are
chosen in a geometric series, and the temporal receptive fields are a function
of $\tau/\taustar_n$, changing $\taustar_n$ by factor of $1/a$ is equivalent to
translating \(n\) such that \(n \rightarrow {n+\Delta}\)
where \(\Delta = \log_{1+c} a\).

With a finite number of nodes, rescaling is not precisely translation.  First, with a
finite number of nodes, information will be lost near the edges of the array.
Second, even neglecting the edges, there is only a precise translation over
the discrete set of nodes if $a$ is chosen such that $\Delta$ is an integer.
However if $c$ is sufficiently small and $k$ is not too big, such that the
blur in the temporal receptive fields is large relative to the spacing between
the nodes, there is a node whose activation will be similar to the initial
node even if $\Delta$ is not an integer.

\begin{figure}[t]
\centering
\includegraphics[width=0.9\columnwidth]{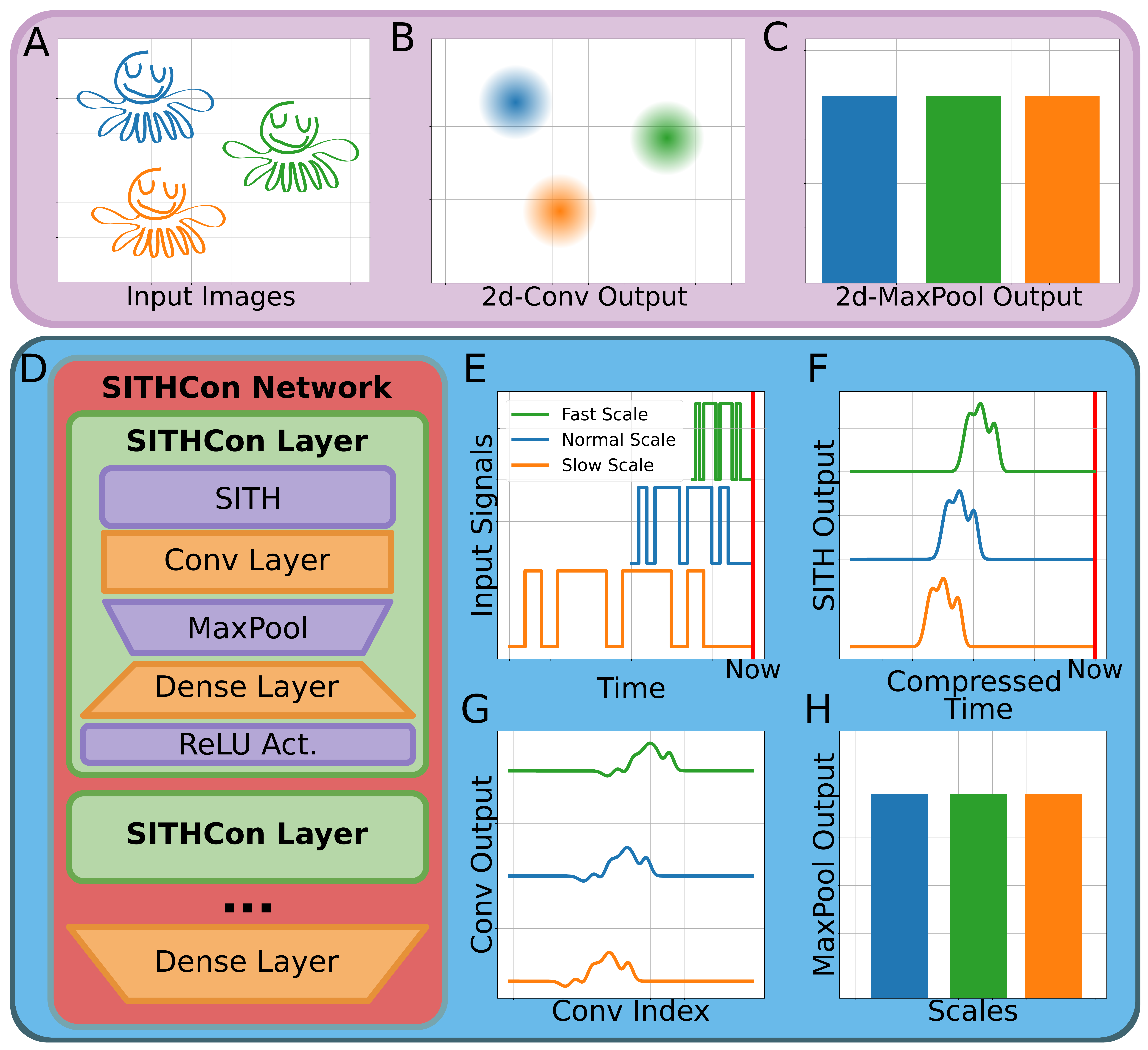}
\caption{\emph{Temporal scale-invariance in SITHCon.} \textbf{A-C:}
Translation-invariance in standard CNNs. \textbf{A:} Example images in
three possible locations. \textbf{B:} 2d convolution output of filters as
a function of position. The output is translated. \textbf{C:} Activation
following 2d max pooling results in translation-invariance. \textbf{D-H:}
Scale-invariance in SITHCon. \textbf{D:} Diagram of the SITHCon network.
Orange represents layers with learnable weights, where purple represents
no learnable weights present. \textbf{E:} A time series \(f(t)\) at three
different time-scales. \textbf{F:} SITH layer output for the different
scales. Because the SITH is logarithmically-compressed, a change
in scale results in translation of the memory. \textbf{G:} Because the
convolutional filters are applied to the output of SITH, the convolutions are
also translated. \textbf{H:} Max pooling the output of the convolutional
layer results in scale-invariance.}\label{sithcon_network}
\end{figure}

\subsection{SITHCon}
\label{sithcon}
The SITHCon network (Fig.~\ref{sithcon_network}.D) is a deep
network.  An external signal with \(\nfeatf\) features provides the input to
SITH at the first layer.   The SITH memory at each layer
at each time point is given by Eq.~\ref{eq:conveqs} operating on the input to
that layer.  We write
\begin{equation}
	\ftilde\layer{i} = \ftildefeat \circ f \layer{i},
	\label{eq:layers0}
\end{equation}
where the operator $\ftildefeat$ is just given by Eq.~\ref{eq:conveqs}, to
describe the SITH memory on the $i$th layer.
The remainder of each layer takes a convolutional neural network over the SITH
memory, followed by a max pooling operation and then a set of dense weights.
The output of the dense weights becomes the input to the next layer:
\begin{eqnarray}
	f\layer{i+1} &=& \sigma\left(
			    W\layer{i} \max_n \left[
				    \convfilt\layer{i} \star
				    \ftilde\layer{i}
                                \right]
                            \right),
        \label{eq:layers}
\end{eqnarray} where \(\convfilt\layer{i}\) are 2-D
convolutional filters of size
\(\nfeatf\times \convwidth\),
\(W\layer{i}\) is a dense layer with
\(\nfeatf\times \nfeatf\) weights and \(\sigma\)
is a ReLU function. The \(\max_n\) operates over the \(\ntaustar\) time indices
rather than the feature indices, thereby returning the maximum output of the
convolution kernel at a specific \(\taustar\) in the past.

The temporal memory in each layer in the network $\ftilde\layer{i}$ provides a
conjunctive representation of what happened when.  Each layer has the same
form of logarithmically-compressed temporal memory.  Critically, however, the
form of the ``what'' information changes from one layer to the next due to the
learned weights from one layer to the next.

Consider how the entire network responds to rescaling the input on the first
layer.  As established in section~\ref{sec:scaletrans}, time rescaling of the
input has the effect of translating $\ftilde$ at the first layer, modulo edge
effects.  The index at which the convolutional filters match this memory will
also translate, again modulo edge effects.  However, the max pool operation
discards information about the index at which the convolution reached its
maximum, so that the features passed on to the next layer at a corresponding
time point are invariant to rescaling. Note, however, that the magnitude of the
rescaling is expressed by the network as the index at which the maximum value
was found.

The region over which edge effects are important is easily calculated.  If the
maximum of a particular filter is found at some index $j$, then if
\(\convwidth/2 < j < \ntaustar-\convwidth/2\) and the rescaling gives $\Delta$
such that \(\convwidth/2 < j + \Delta < \ntaustar - \convwidth/2\), then edge
effects can be neglected.
In practice, different features at different layers will not in general have
the same maximum index, so with enough features and enough layers, there is a
real possibility that the network as a whole is not robust to time rescaling
even if each of the layers taken individually is invariant over a wide range.
We treat it as an empirical question whether these ranges overlap for the
problems studied here.

These considerations suggest a strategy to develop networks that
rescale over an arbitrarily large range of scales without retraining.  After
training a network on a particular problem, one can simply add $\taustar$
nodes to each layer of the network.  Even if the original network does not
generalize over scales, one is guaranteed that all of the features learned at
all of the layers will be able to scale over the added nodes.  The range of
scales over which the network will generalize should go up
exponentially like $\left(1+c\right)^N$, where $N$ is the number of
additional nodes.  Notably, the number of weights is unaffected by the number
of nodes added to the network.   The convolution simply operates over a larger
range.  In this way, one ought to be able to construct efficient SITHCon
networks that generalize over an exponentially wide range of scales with
no additional commitment of training time.

\subsubsection{Relation to Previous Work}

SITHCon is closely related to the  DeepSITH network \citet{Jacques.etal.2021}.
Both networks utilize SITH as a compressed
representation of the past. Both networks include dense connections from one
layer to the next.  The core difference between these two models is
that SITHCon, but not DeepSITH, includes a convolutional layer and maxpool
operation at each layer.

DeepSITH was applied to a variety of challenging time series problems.  Unlike
RNNs, including LSTMs, DeepSITH was able to learn time series problems even
when they required the network to learn very long-range dependencies.
Section~\ref{sec:scaletrans} provides some insight into why DeepSITH is able
to learn time series problems with very long-range dependencies.  DeepSITH
weights take the representation of compressed time  of size $(\taustar,
\nfeat_{i})$ to the inputs at the next layer $(\nfeat_{i+1})$.  Consider how
DeepSITH would behave if, after training on a problem with some time series
$f(t)$, the network was trained on the same problem with time rescaled  $t
\rightarrow at$.  Because of the temporal memory at
each layer responds to time rescaling as translation along the log time axis,
we know that DeepSITH would provide the same output to $f(t)$---neglecting
edge effects---if all of the weights were translated along the time index by
$\Delta=\log_{1+c} a$.  In this sense, DeepSITH's ability to learn
time series problems is invariant to the time scale of the problem.

The addition of convolution and maxpooling enables SITHCon to respond
equivalently---neglecting edge effects---to time-rescaled inputs \emph{without
retraining}.
This also means that the two networks would behave differently to a training
set that includes problems at many time scales.  Whereas SITHCon is able to
use the same weights for $f(t)$ and $f(at)$, DeepSITH would require additional
weights to learn the different scales.  For this reason, we would expect DeepSITH
to learn problems with a mixture of scales across training examples more
slowly and at greater cost in terms of number of weights than SITHCon.

\subsection{TCN}
\label{tcn}

In the following experiments, we compare SITHCon to the TCN
from \citet{Bai.etal.2018}. The TCN was introduced as a
generic convolutional architecture used in sequence modeling tasks, and
is well regarded as a replacement for canonical recurrent neural
networks (RNN). This is due in part to its performance on many machine
learning benchmarks that require a long temporal history. SITHCon and
TCN are similar in that both are made up of a series of layers that
encode information at various scales. TCN networks use causal
convolutions, preventing any leakage of future information into the
past. In addition, these convolutions have exponentially increasing
dilations, which gives each layer an ``effective history'' of
\((\convwidth- 1)\) times the  dilation.

There are two fundamental differences between TCN and SITHCon. The first
is that TCN convolutions operate directly on normal time, where SITHCon
layers apply their convolutions to compressed time. The second is that a
SITHCon network's effective history is limited by their largest
\(\taustar\), which goes up exponentially with the number of nodes. The effective
history of the TCN is limited by the number of layers and the dilation at each
layer.

We use the TCN implementation supplied by \citet{Bai.etal.2018} at
https://github.com/locuslab/TCN. Each TCN has eight layers with 25 channels
each. The kernel size was chosen for each experiment to give a
reasonable number of weights.

\section{Experiments}
\label{experiments}

We examined the performance of SITHCon and TCN on two classification
tasks, and one regression task. In each experiment, the networks were
trained at a single training scale (or a few scales in Exp. 4). After the networks
were fully trained, we then compared SITHCon and TCN in their ability to
generalize to unseen scales.

\begin{table*}
\begin{center}
\begin{sc}
\begin{small}
\begin{tabular}{ l | c c c c | c c c c }
& \multicolumn{4}{c}{\underline{SITHCon}} & \multicolumn{4}{c}{\underline{TCN}}\\
Exp & Wts. & Layers & $K$ & Chan. & Wts. & Layers & $K$ & Chan. \\
\hline
1. Morse Decoder & 33k & 2 & 23 & 35 & 142k & 8 & 14 & 25 \\
2. Morse Addition & 31k & 2 & 23 & 25 & 425k & 8 & 46 & 25 \\
3. AudioMNIST & 71k & 2 & 23 & 35 & 171k & 8 & 16 & 25 \\
\hline
\end{tabular}
\end{small}
\end{sc}
\caption{\emph{Network parameters for each experiment.} $K$ is the kernel size for the convolutions. Chan. is the number of convolution channels.
 \label{tab:model_params}
}
\end{center}

\end{table*}

The Morse Decoder task requires the networks to classify the 43 Morse
code digits presented as a time series. The Morse Addition task uses two
input features and is very similar to the Adding Problem benchmark. The networks
must learn to add the values of two Morse
code digits presented within a continuous stream of digits, and marked
by active bits in a parallel time series. This task requires both digit
recognition and memory, which must both be maintained with changes in
temporal scale. Finally, the Audio MNIST task \cite{Becker.etal.2019} requires the
networks to classify spoken digits 0-9 in recordings by many different
speakers. We ran each experiment five times for each network with different
seeds to get a measure of variability in performance.

In all experiments, SITHCon was similarly configured, with two SITHCon
layers, each with 400 values of \(\taustar\)
log-spaced from 1 to 3000 or 4000 and
\(k\) of 35. The width of the convolution kernels was
set to 23 with a dilation of 2. The only difference in the model between
the experiments was the number of convolution channels, which were
varied based on the complexity of the problem. The TCN was also largely
similar across experiments, with 8 total layers, only varied in number
of input channels and the kernel width. We list the comparable
parameters between the TCN and SITHCon networks in
Table~\ref{tab:model_params}.

\begin{figure}
\centering
\includegraphics[width=0.49\textwidth]{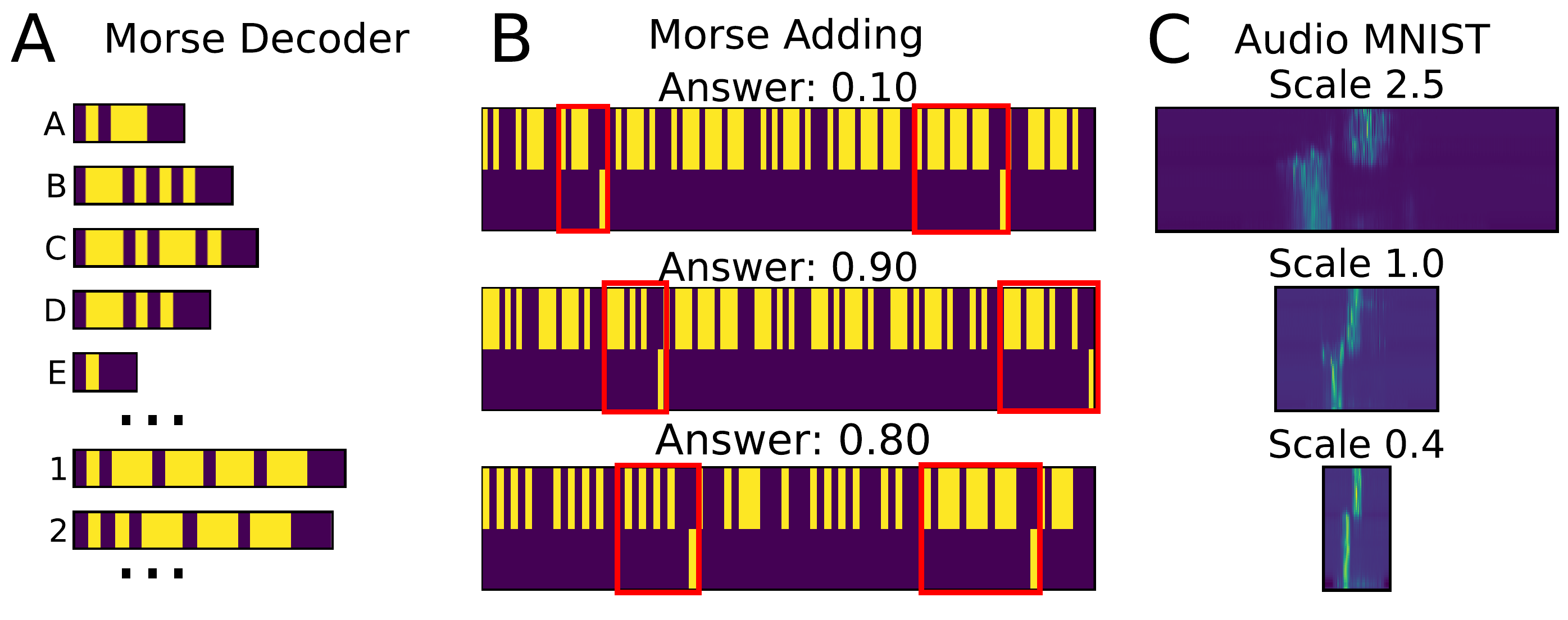}
\caption{\emph{Stimuli and tasks for the three experiments.} Time is
on the x axis. \textbf{A}: The Morse Decoder
problem requires the network to learn the label associated with the 43
Morse code symbols (seven of which are shown here). Yellow is an ``on'' bit,
while blue is an ``off'' bit. \textbf{B}: Morse Addition takes
two-dimensional time series as inputs. One dimension contains a stream of ten
Morse code digits, the other contains two activation pulses
indicating which Morse code symbols are to be added (shown in red). The
correct answer for the network is 0.1 times the sum of the two indicated
digits (shown above each example time series). \textbf{C}: The Audio
MNIST task requires the network to recognize the label associated with
spoken digits from a variety of speakers. Here a single clip (the
spoken word \textsc{seven}) is shown as a spectrogram (normalized power as
function of frequency and time) at three different scales. In Exp.
3.A the networks learned stimuli at scale 1.0, and were tested at the other
scales. In Exp. 3.B the networks are trained on five scales and
then tested on stimuli from those five and four additional interleaved
scales.}\label{exp_examples}
\end{figure}

\subsection{Exp. 1: Morse Decoder}
\label{exp.-1-morse-decoder}

Morse code is a standardized way to encode text into a one-dimensional
time series comprising different sequences of dots and dashes (i.e.,
short and long activation periods), each separated by short periods of
silence. Differentiating the 43 different Morse code signals is a
relatively simple time-series classification problem because each Morse
code symbol is a unique pattern of dots and dashes.
We trained SITHCon and a TCN to differentiate
the Morse code symbols at a single scale. Then we tested the two
networks on Morse code symbols presented at a range of unseen temporal
scales.

\subsubsection{Exp. 1.A: Signal classification at multiple scales}
\label{exp.-1.a-signal-classification-at-multiple-scales}

The training dataset consisted of 43 Morse Code symbols (letters,
numbers, and punctuation/symbols). Each dot in a symbol was represented
by the signal being ``on''--set to a value of 1---for one time step and
``off''---set to to 0 for one
time step. Each dash was represented by the signal being ``on'' for three
time steps and ``off'' for one time step. At the end of each symbol, the last dot or
dash was followed by three time steps of ``off''. Examples of these symbols
are shown in Fig. \ref{exp_examples}.A. We trained these
networks with the symbols where each time step was of length 10.  This scale
was  designated as scale 1.0, so that we could easily scale both up and
down relative to the training sequences.

\begin{figure*}
\centering
\includegraphics[width=0.9\textwidth]{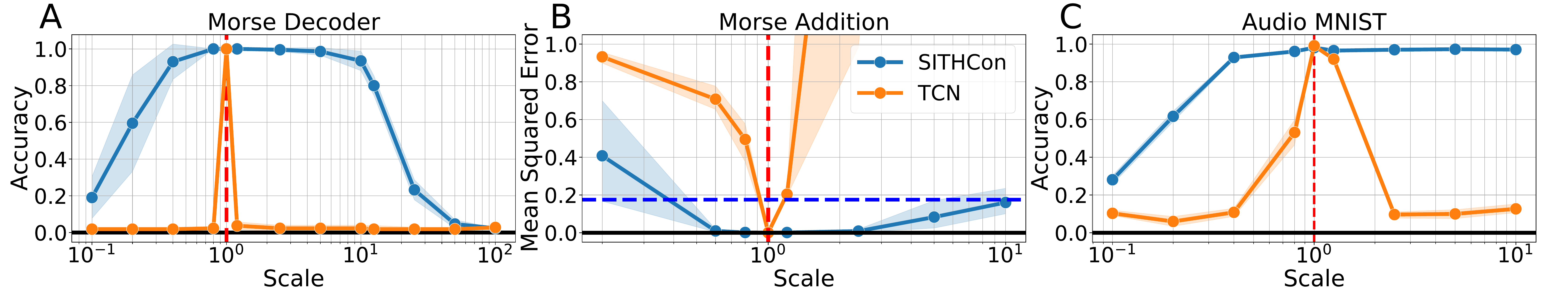}
\caption{\emph{SITHCon generalizes over unseen scales.} Performance on
Exp. 1.A, 2, and 3.A as a function of testing scale.
The dashed vertical red line in each plot indicates the training scale.
Error bars are 95\% confidence intervals over 5 distinct runs.
\textbf{A}: Morse Decoder, performance measured with accuracy on Morse
code digits scaled to different lengths. \textbf{B}: Morse Addition,
performance measured with mean squared error on a held out test set of
1000 scaled signals. \textbf{C}: Audio MNIST, a classification task
involving multiple recordings of spoken digits preprocessed with wavelet
decomposition. Performance measured with accuracy on held out
recordings, scaled in tempo for each test
scale.}\label{performance_full}

\end{figure*}

Once the networks reached 100\% classification accuracy on the 43 Morse
code symbols, we tested the networks on various  time-rescalings using
the same trained weights. We take each of the Morse Code signals and repeat
every bit different numbers of times. For example, to test a network's accuracy
at 2 times the training scale, we repeated every bit 20 times.
Fig. \ref{performance_full}.A shows the results of SITHCon and
TCN on different scales. As expected, TCN and SITHCon were able to reach
100\% accuracy at a scale 1.0. However, the two networks showed very
different generalization across scales. Whereas the TCN fell to
chance-level performance with even a small variation from the training
scale, SITHCon showed close-to-perfect generalization over scale
increases of an order of magnitude and was above chance over more than
two orders of magnitude.

\subsubsection{Exp. 1.B: Maximum effective range increases exponentially with number of added $\taustar$ nodes}
\label{exp.-1.b-toward-examining-the-maximum-effective-range-of-sithcon}

\begin{figure}
\centering
\includegraphics[width=0.9\columnwidth]{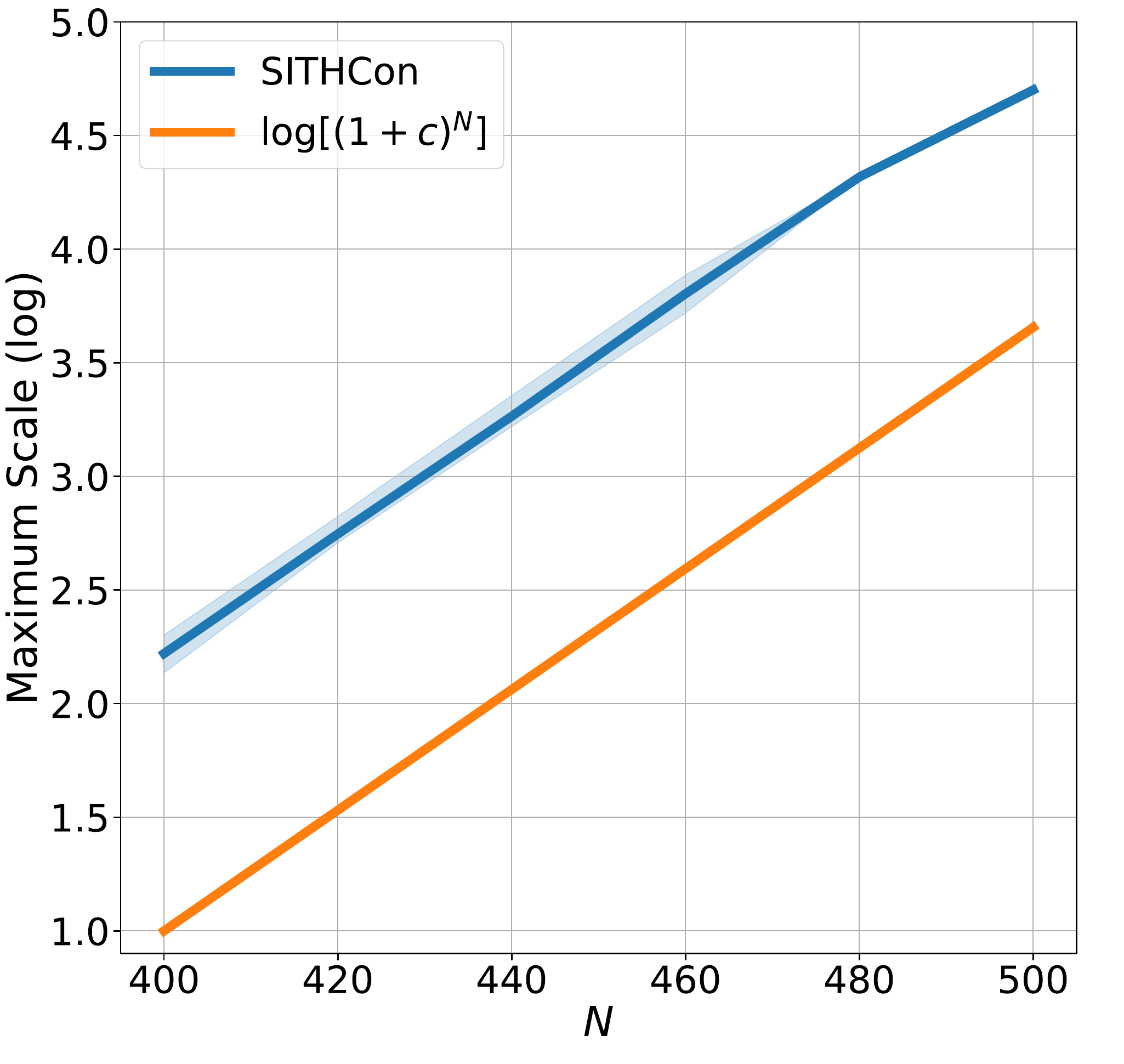}
\caption{\emph{The largest effective scale grows exponentially with the number
of $\taustar$ nodes added to the network.}
After training the SITHCon network on the Morse Code classification task with
$N_{\taustar}=400$ we added $\taustar$ nodes to the already-trainined
network.   We measured the range of scales over which the network successfuly
generalized (y axis, on a log scale) as a function of the number of nodes (x axis).   The
orange line shows the theoretical curve given by $\left(1+c\right)^N_{\taustar}$.
The effective maximum scale increases
exponentially. }\label{eff_max}

\end{figure}

Considerations discussed earlier in section \ref{sec:scaletrans} suggest
that the effective maximum time-rescaling of an already-trained SITHCon
network should increase like
\((1+c)^{\ntaustar}\) as nodes are added. We define maximum time-rescaling for
this experiment to be the maximum amount we can temporally scale the
training signals while still achieving 100\% accuracy. This experiment took a SITHCon network
trained to classify Morse Code digits at one
scale using the procedure in Exp. 1.A,
and simply changed \(\ntaustar\).  Critically, the nodes were
added to the trained network with the same relative spacing.
Fig. \ref{eff_max} shows that the range of scales over which the network
generalizes goes up exponentially like \((1+c)^{\ntaustar}\).
This property should hold for any SITHCon network trained on any task,
enabling generalization over exponentially large scales with a linear increase
in memory and no increase in the number of trained weights.  Of course, this
procedure is only helpful if the initially-trained network can learn the
problem at hand.

\subsection{Exp. 2: Morse Addition}
\label{exp.-2-morse-adding}

For this experiment we developed a novel variant of the Adding Problem
\cite{Hochreiter.Schmidhuber.1997}, which we refer to as the Morse Addition. In this task,
SITHCon and TCN networks received a two-dimensional time-series input.
As illustrated in Fig. \ref{exp_examples}.B, the first
dimension was composed of a continuous stream of ten Morse code symbols,
which we mapped onto the numbers 0.0 through 0.9 and selected at random
and with replacement to form a sequence representing numerical values.
The second dimension was only zeros except for two pseudo-randomly
selected locations with a value of 1.0, one occurring in the first half of
the signal and one occurring in the latter half. The goal of the task
was to decode the Morse code symbols indicated by the bits in the second
dimension, and then add the two decoded symbol values together.
For example, if the symbols for 0.1 and 0.6 were identified as the
targets within the stream of 10 Morse code symbols, the networks would
have to output 0.7.

We trained the TCN and SITHCon networks on the Morse
Adding Problem such that each bit in the sequence was repeated five times at scale 1.0. We trained both networks to
minimize Mean Squared Error (MSE). Once trained, we
evaluated each network's ability to perform the task with the input sequences at
other scales. The TCN had 425k weights, and SITHCon had 31k.

The results in Fig. \ref{performance_full}.B demonstrate that, as
expected, both the TCN and SITHCon showed perfect performance at the
training scale. The dashed blue line in Fig.
\ref{performance_full}.B represents chance performance of a
hypothetical network that simply guessed the mean of possible target
values on every trial. As the testing scale deviated from the training
scale, TCN suffered from rapid deterioration in performance, showing
performance worse than the hypothetical network when the signals were
rescaled even by a single bit. At large scales, the TCN performed very
poorly (Note the y-axis is truncated). In contrast, the SITHCon
network maintained low error rates for changes in scale
between .6 and 2.4, and remained better over changes in scale of about an order of
magnitude.  As discussed above, adding nodes to the already-trained network
would result in an exponentially large range of generalization without
learning additional weights.

\subsection{Exp. 3: Auditory MNIST}
\label{exp.-3-auditory-mnist}

The AudioMNIST dataset consists of 60 speakers, $33\%$ female,
who were recorded speaking individual digits (0-9) 50 times each
\cite{Becker.etal.2019}. We used this collection of audio clips to create a training
dataset consisting of 45 out of 50 stimuli for each digit from all
speakers. The remaining 5 stimuli per digit from each speaker were used
for testing. We first padded each 48kHz clip to 50,000 samples with an
equal number of zeros at the front and back of the recordings. Then the
stimuli were passed through a Morlet wavelet transform with 50 log-spaced
frequencies from 1000Hz to 24kHz and then Z-scored within frequency
across time and downsampled to 240Hz. The resulting stimuli for this
task had 50 input features and 250 time points at the standard training
scale.

To test the scalability of the TCN and SITHCon networks, we created
stretched and compressed versions of each audio clip via pitch-locked
time-scale modification prior to the wavelet transformation \cite{Driedger.Muller.2016, Muges.2021}
(see Fig.~\ref{exp_examples}.C for example spectrograms of scaled
stimuli). These scales ranged from ten times slower (scale $> 10^0$)  to ten times
faster (scale $< 10^0$).

After training on scale 1.0 for ten epochs, the networks were tested on
held-out stimuli from a range of scales. The results of this test are
shown in Fig. \ref{performance_full}.C. Both networks showed
essentially perfect performance when the test scale was the same as the
training scale. The TCN showed some generalization for small variations
in the testing scale, but fell to chance performance rapidly. In
contrast, SITHCon maintained a high level of testing accuracy over
scales that varied from the training scale by more than an order of
magnitude. As discussed above, adding nodes to the already-trained network
would result in an exponentially large range of generalization without
learning
additional weights.

\begin{figure*}
\centering
\includegraphics[width=0.8\textwidth]{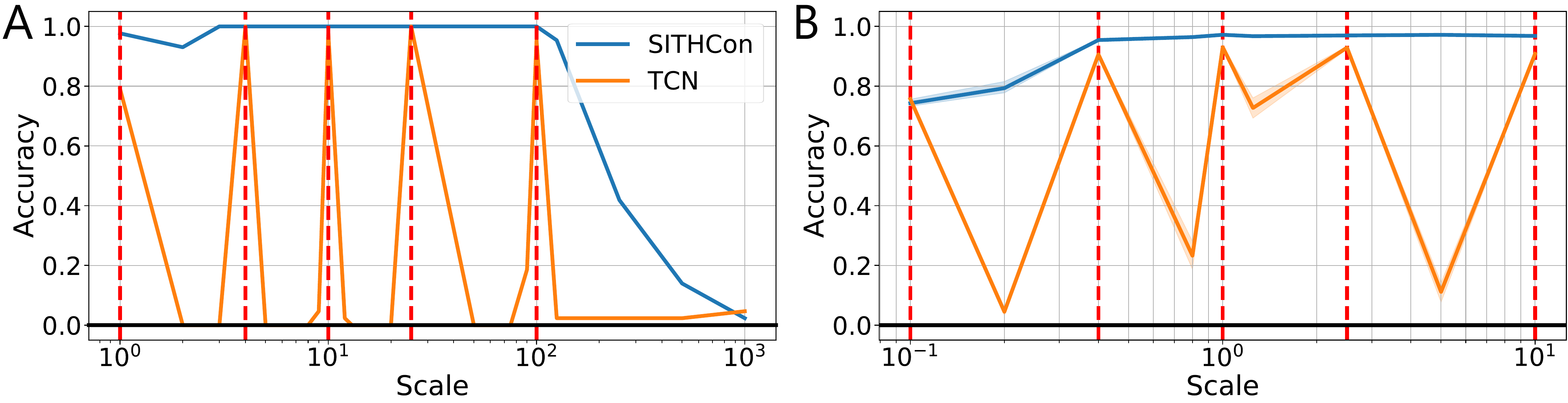}
\caption{\emph{Exp. 4, training at multiple scales and testing for
	generalization to intermediate scales.} Here, Exp. ~1~ and Exp. ~3~ are repeated with
additional scales included during training. We trained networks on scales .1,
.4, 2.5, 10.0, in addition to the standard training scale of 1.0. \textbf{A}:
	Repetition of Exp.~1. Each network was trained with the entire training set at every training
scale, effectively creating a five-times larger training set. As expected,
SITHCon was able to scale at intermediate scales, while TCN still was not able
to generalize well to intermediate scales.
\textbf{B}: Repetition of Exp.~3. Unlike in \emph{A}, the total number of
training samples was the same as in Exp.~3, and one fifth of the samples were
scaled to each training time-scale. SITHCon generalized well across untrained
scales. The TCN did not generalize well to intermediate scales, and performed
worse than in Exp.~3 on the training scales. }
\label{audio_mnist_mixed}
\end{figure*}

\subsection{Exp. 4: Variable Scale Training}
\label{exp.-4-variable-scale-training}

Above, we have shown that while training on a single scale, the TCN is unable
to generalize to unseen scales. In this experiment, we re-examine
Exp. 1 and Exp. 3 as Exp. 4.A and 4.B respectively. Rather than training only on
scale 1.0, the networks were trained on scales .1, .4, 2.5, and 10.0.

In Exp. 4.A, the amount of training items is 5x larger
than in Exp. 1, as we time-rescaled all of the training items to all five
training scales before the experiment. In Exp. 4.B, we
decided to keep the total number of training items the same as in Exp. 3. Each
run, we would randomly select a fifth of the training items to always be scaled
to .1, a fifth to .4 scale, etc.

The results are shown in Fig. \ref{audio_mnist_mixed}. For both experiments we
see that only the SITHCon network could generalize to unseen scales.
Fig. \ref{audio_mnist_mixed}.A shows that the TCN learned to identify
Morse code signals at the training time-scales in Exp. 4.A, but was unable to
generalize. Meanwhile SITHCon was able to generalize better to the unseen faster
scales that it was unable to scale to previously. Fig.
\ref{audio_mnist_mixed}.B shows SITHCon well outperforms
the TCN at the training time-scales, as well as the unseen time-scales. This is
likely due to the fact that SITHCon treats differently time-scaled
training samples of the same word as similar, whereas the TCN has to learn
how to recognize each different time-scale and word individually.
SITHCon is able to treat differently time-scaled time-series signals
as similar, and therefore save computational resources by not having to learn to
identify them separately.

\section{Discussion}
\label{discussion}

This paper presented SITHCon, a deep convolutional neural network built from
layers of logarithmically-compressed, scale-invariant representations of the
recent past and pooled convolutions. Rescaling the input signal in time
results in a translation of the state of SITH over indices. Because
the output of the CNN layer depends on the maximum of the activity over
indices, each convolutional filter is scale-invariant over a wide
range of time-rescalings, limited only by edge effects.
\citet{Jansson.Lindeberg.2021} built a visual CNN that exploits similar ideas
to generalize to visual images of different sizes.

We performed a series of experiments and found that SITHCon generalized
to a wide range of unseen temporal scales; SITHCon generalized over time
rescalings of about an order of magnitude.  TCN did not generalize to time
rescalings of the input signal.  Techniques used for speech recognition prior
to the rise of deep networks, were robust to changes in the rate of speech
\citep[e.g.,][]{SakoChib78}.
Although deep networks have replaced these
techniques, their memory representations require training on a variety of speech
rates. Deep convolutional networks with a
logarithmically-compressed temporal memory provide a strategy that could
combine the power of deep networks with a human-like ability to generalize to
time-rescaled input signals.

The SITHCon network also has capabilities that are very different from natural
learners.  To obtain arbitrarily large ranges of time-rescaling, one can take
a trained network and simply add \(\taustar\)'s to enable the network to
generalize over whatever range of scales is desired.  The range of scales over
which the network can generalize goes up exponentially with the number of
\(\taustar\)'s added to the network with no additional learned weights
(Fig.~\ref{eff_max}).

The strategy employed in SITHCon would work for any network with
logarithmically-compressed temporal basis functions and retains
indices of the basis functions in an organized way to enable invariance
\cite{LindFage96, deVrPrin92}. It should be noted that general RNNs are
extremely ill-suited for this purpose. It is possible to write the
set of linear filters in Eqs.~\ref{eq:geometric}~and~\ref{eq:conveqs} as a
recurrent network \cite{Liu.Howard.2019}---resulting in a scale-covariant RNN.
However, it is not at all obvious how to ensure that the RNN would reach that
state after training without introducing constraints very similar to SITH.
Moreover, even if an RNN generated a set of temporal basis functions, it would
not be clear how to access the indices of the temporal basis functions in order
to make the memory scale-invariant.


In this study, we computed \(\ftilde_t\) by
directly convolving the input signal with \(\phi\), which requires
retaining the entire signal \(f_t(\tau)\) at each moment in time.
Insofar as \(\ftilde_t\) samples the signal at
geometrically-spaced points (Eq.~\ref{eq:geometric}), one could save
memory if it were possible to update \(\ftilde_t\)
without retaining each \(f_t\). One possibility is to compute an
estimate of the real Laplace transform of \(f_t\), \(F_t(s_n)\) with
\(s_n=k/\taustar_n\). Each of these nodes in \(F_(s_n)\) can be updated
using only the value of \(f\) at that moment and the node's previous state.
One can estimate \(\ftilde_t\) by using the Post
approximation to the inverse Laplace transform, which requires taking
the \(k\)th derivative with respect to \(s\). In practice, the Post
approximation becomes numerically unstable for high values of \(k\). A
related approach is to construct \(\ftilde_t\)
from the \(F_t(s_n)\) using a cascade of convolutions of \(F(s_n)\)
values \cite{Lindeberg.2016}. The gamma network offers another solution \cite{deVrPrin92}.

For a scale-invariant network, training
examples at different speeds do not interfere with one another as they
are treated identically by the network. Many natural signals contain
information at very different temporal scales. In practice, machine
learning applications have often addressed this problem by brute force,
training the network on many different examples \citep[e.g.,][]{Chan.etal.2021}, as in
Exp.~4. Perhaps the brain has evolved logarithmically-compressed
temporal basis functions \cite{Guo.etal.2020} to endow it
with the ability to speed up learning and rapidly generalize to unseen
experiences.

In vision, researchers have long appreciated the importance of
generating features that are invariant to time-rescaling of the input
\cite{Lowe.1999}. Jansson and Lindeberg \citeyear{Jansson.Lindeberg.2021} use an approach similar to that
used here for MNIST digits of various sizes. By generating a set of scaled
representations of visual images and integrating features over
logarithmically-spaced scales they achieve scale-invariance over a broad
range of spatial scales for essentially the same reason that we observed
effective scale-invariance over a range of temporal scales in this paper
\citep[see also][]{Lindeberg.2016}. In natural vision, the image on the retina is not
constant.  Even when objects in the world are prefectly still, movement of the
eyes still induces rich dynamics over a range of temporal and spatial scales.
Rather than images, natural vision operates on spatiotemporal patterns \cite{Rucci.Victor.2015}.
Understanding how scale-covariance in both time and space could be
used to inform computer vision open question of considerable theoretical and
practical importance.

The Weber-Fechner Law is widely observed in behavior across mammals
suggesting that the strategy of logarithmically-compressed basis
functions seems to be used quite broadly in the brain. In many cases,
such as perception of time or nonverbal numerosity \cite{Gallistel.Gelman.2000, Dehaene.Brannon.2011}, the
logarithmic distribution cannot be attributed to the physical structure
of a sensory organ. This suggests the possibility that learning rules
attempt to map representations onto continuous logarithmically-compressed
dimensions. Such representational spaces could naturally support the kind of
scale-invariance illustrated in this paper and allow for powerful information
processing with a simple neural architecture.

\bibliography{SITHCon,citesforicml}
\bibliographystyle{icml2022}

\end{document}